%% file: main.tex
\documentclass[10pt,twocolumn,letterpaper]{article}

\usepackage{wacv}              

\input{preamble}

\definecolor{wacvblue}{rgb}{0.21,0.49,0.74}
\usepackage[pagebackref,breaklinks,colorlinks,allcolors=wacvblue]{hyperref}

\def\wacvPaperID{1891} 
\def\confName{WACV}
\def\confYear{2026}

\title{MedPEFT-CL: Dual-Phase Parameter-Efficient Continual Learning with Medical Semantic Adapter and Bidirectional Memory Consolidation}

\author{
Ziyuan Gao\\
University College London\\
}

\begin{document}
\maketitle
\begin{abstract}
Medical vision-language segmentation models suffer from catastrophic forgetting when adapting to new anatomical structures, requiring complete retraining that limits their clinical deployment. 
Although continual learning approaches have been studied for various applications, targeted research on continual learning approaches specifically designed for medical vision-language tasks remains underexplored.
We propose MedPEFT-CL, a parameter-efficient continual learning framework that addresses both efficient learning of new tasks and preservation of previous knowledge through a dual-phase architecture based on CLIPSeg.
Our dual-phase architecture features an adaptive learning phase that employs semantic similarity-based adapter allocation and parameter-efficient fine-tuning for medical tasks through prompt similarity analysis, and a knowledge consolidation phase employing bi-directional Fisher-memory coordination. 
This creates a reinforcing cycle: consolidation directs replay priorities while new tasks provide challenging samples that improve retention strategies.
Our key contributions are: (1) a semantic-driven adapter allocation mechanism that enables efficient learning of new medical tasks, (2) a bi-modal LoRA adaptation that significantly reduces trainable parameters while maintaining cross-modal learning, and (3) bidirectional Fisher-memory coordination that prevents catastrophic forgetting from previous medical tasks.
Extensive experiments across diverse medical datasets demonstrate superior forgetting mitigation and performance retention with minimal parameter overhead, making the framework effective for continual learning in medical vision-language scenarios.
\end{abstract}

\section{Introduction}
\label{sec:introduction}

\begin{figure}[h]
\centering
\includegraphics[width=\columnwidth]{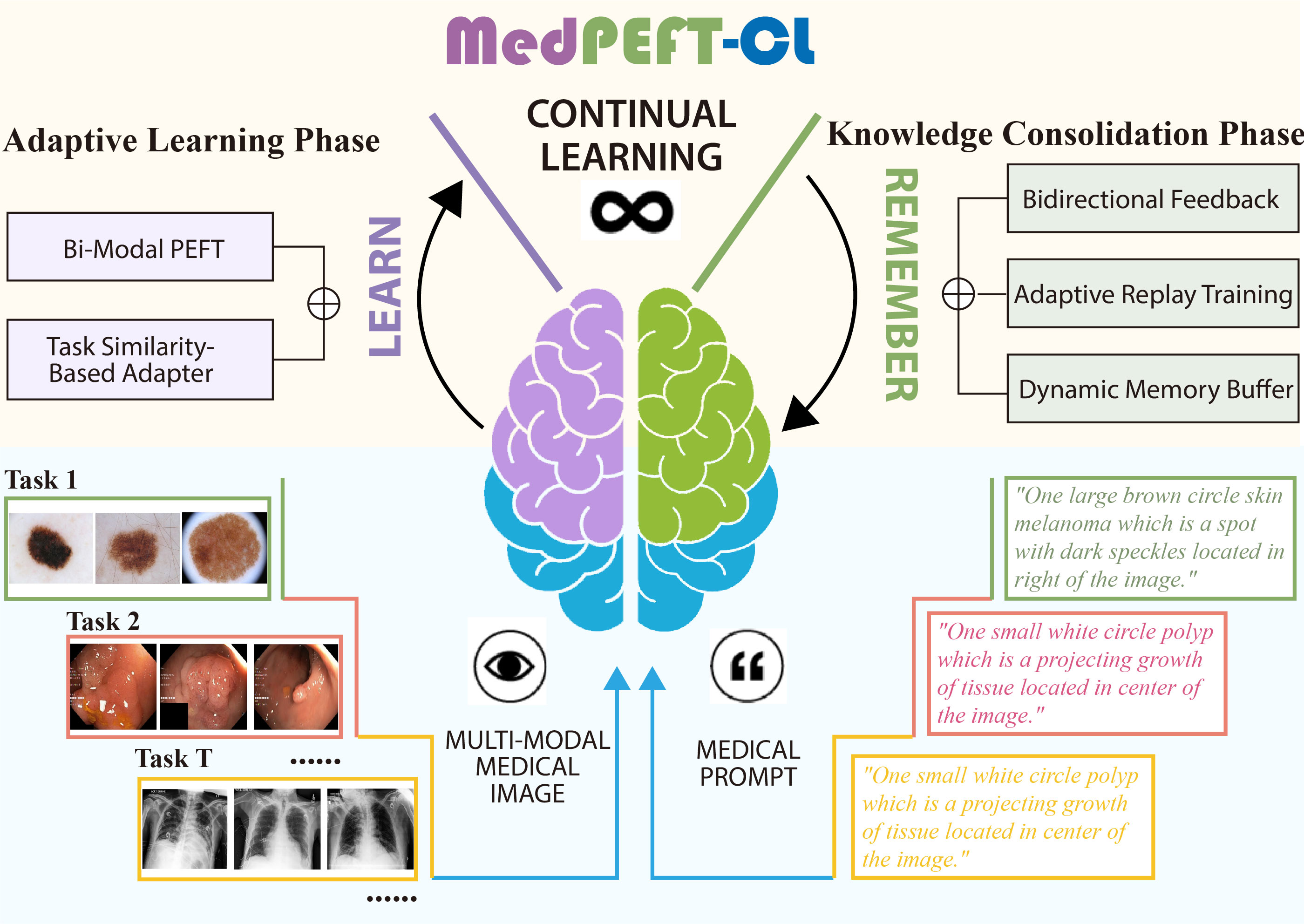} 
\caption{We propose \textbf{MedPEFT-CL}, a PEFT framework that addresses both efficient learning of new medical tasks and preservation of previous knowledge through a dual-phase architecture for medical visual-language segmentation model.}
\label{fig:pipeline}
\end{figure}

Medical AI systems face multiple unique deployment challenges including complex workflow demands for seamless integration into existing clinical practices~\cite{davenport2019potential,kelly2019barriers}, significant domain shifts when deploying across different healthcare institutions~\cite{qazi2024continual,thandiackal2024multiscale}, and the need for continuous adaptation to evolving clinical 
protocols~\cite{bruno2024continual,zhao2023class}. Unlike conventional AI domains, medical applications demand both exceptional performance retention and strict computational efficiency due to resource constraints in clinical environments~\cite{bayasi2024gc2,gao2023incremental}. These challenges make continual learning particularly critical for medical applications.

When adapting general vision-language models like CLIPSeg to medical domains, sequential learning of new anatomical structures causes catastrophic forgetting as new task parameters overwrite previously learned knowledge~\cite{hofmanninger2021dynamic,zhai2023investigating}. This phenomenon is particularly problematic in medical settings where both visual and textual representations must be preserved across diverse anatomical tasks~\cite{luo2023empirical}, forcing practitioners to retrain models from scratch for each new application rather than building upon existing knowledge. While continual learning approaches have been extensively studied for computer vision and natural language processing tasks, targeted research on continual learning frameworks specifically designed for medical vision-language applications remain significantly underexplored. Existing approaches either focus on single-modality medical tasks~\cite{bayasi2024gc2,liao2022muscle} or general-domain vision-language models~\cite{VR-LwF2022,MoE-Adapter2024}, leaving a critical gap in addressing the challenges of medical cross-modal learning scenarios.

The medical domain presents distinct challenges for continual learning~\cite{hofmanninger2021dynamic}. Medical models must handle diverse imaging modalities and anatomical structures while maintaining diagnostic accuracy across previously learned tasks~\cite{salahuddin2023redefining}. Additionally, clinical deployment constraints require parameter-efficient solutions that can adapt to new medical domains without extensive computational overhead~\cite{davenport2019potential}. These factors necessitate specialized continual learning approaches that address the unique requirements of medical vision-language applications.

We propose \textbf{MedPEFT-CL}, a parameter-efficient continual learning framework that addresses both efficient learning of new medical tasks and preservation of previous knowledge through a dual-phase architecture, as shown in Figure~\ref{fig:pipeline}. Our approach employs semantic similarity-based adapter allocation and bidirectional Fisher-memory coordination. This creates a reinforcing mechanism where consolidation directs replay priorities, while new tasks provide challenging samples that improve retention strategies.

Our key contributions are: (1) \textbf{Semantic-driven adapter allocation} that leverages medical prompt similarity to enable efficient parameter sharing across related anatomical segmentation tasks; (2) \textbf{Bi-modal LoRA adaptation} that significantly reduces trainable parameters (0.24-0.39M vs 150M) while maintaining cross-modal learning capabilities; (3) \textbf{Bidirectional Fisher-memory coordination} that creates a reinforcement loop between importance weighting and memory replay to prevent catastrophic forgetting; and (4) \textbf{Comprehensive evaluation} across diverse medical datasets spanning endoscopy, dermatology, ultrasound, and radiology demonstrating superior forgetting mitigation (1.91\% vs 6.21\% for best baseline) and performance retention with minimal parameter overhead. This approach makes continual learning practically viable for clinical deployment while maintaining diagnostic accuracy across evolving medical tasks.


\section{Related Work}
\label{sec:Related Work}

\noindent\textbf{Continual Learning.}
Traditional continual learning methods like EWC~\cite{Kirkpatrick2017} and LwF~\cite{Li2017} use Fisher information weighting and knowledge distillation to prevent catastrophic forgetting in single-modality tasks. DER~\cite{Buzzega2021} combines rehearsal with knowledge distillation for improved memory replay. PackNet~\cite{mallya2018packnet} uses parameter isolation by learning binary masks for each task, while Progressive Neural Networks~\cite{rusu2016progressive} allocate dedicated parameters for new tasks. Recent work has extended these approaches to vision-language scenarios. VR-LwF~\cite{VR-LwF2022} adapts LwF for CLIP-based models. MoE-Adapters~\cite{MoE-Adapter2024} introduces mixture-of-experts mechanisms for continual learning. However, research on domain-specific adaptations remains underexplored across specialized fields, particularly in healthcare applications.

\noindent\textbf{Parameter-Efficient Fine-Tuning in Medical Applications.}
PEFT methods like LoRA~\cite{hu2021lora} enable efficient adaptation by fine-tuning only a small subset of parameters while freezing pretrained weights. PeFoMed~\cite{li2024pefomed} applies PEFT techniques to medical VQA using LoRA adaptations on multimodal foundation models. Low-rank MoE combines mixture-of-experts with low-rank decomposition for enhanced parameter efficiency across multiple tasks~\cite{Che_LowRank_MICCAI2024}. Recent comprehensive analyses demonstrate PEFT's effectiveness across medical domains, with performance gains up to 22\% in medical text-to-image generation tasks while significantly reducing computational costs~\cite{sharma2024peft}. Specialized medical applications include continual classification of medical images~\cite{bayasi2024gc2}, multi-task self-supervised learning for X-ray analysis~\cite{liao2022muscle}, and incremental learning from evolving medical ultrasound streams~\cite{gao2023incremental}. Despite these advances, the integration of PEFT with continual learning paradigms remains  unexplored, especially for medical applications.

\noindent\textbf{Medical Vision-Language Foundation Models.}
Recent medical VLMs have shown significant progress. MedVLSM~\cite{moon2024medvlsm} provides comprehensive medical prompts with clinical contexts. MedCLIP~\cite{wang2022medclip} offers contrastive learning from unpaired medical data. ClinicalBLIP~\cite{zhang2023clinicalblip} demonstrates automated report generation capabilities. Various specialized applications span from cross-modal alignment to organ segmentation and tumor detection. However, these models are static without anti-forgetting mechanisms and require retraining from scratch for each new task, creating computational barriers for clinical deployment. Adapter-based approaches in medical settings face unique constraints including strict memory limitations in clinical hardware, the need for rapid task switching between different medical specialties, and privacy-preserving requirements that limit traditional rehearsal methods~\cite{houlsby2019adapters,li2021prefix,thandiackal2024multiscale,zhao2023class}.
\section{Method}
\label{sec:Method}

\begin{figure*}[t]
\centering
\includegraphics[width=\textwidth]{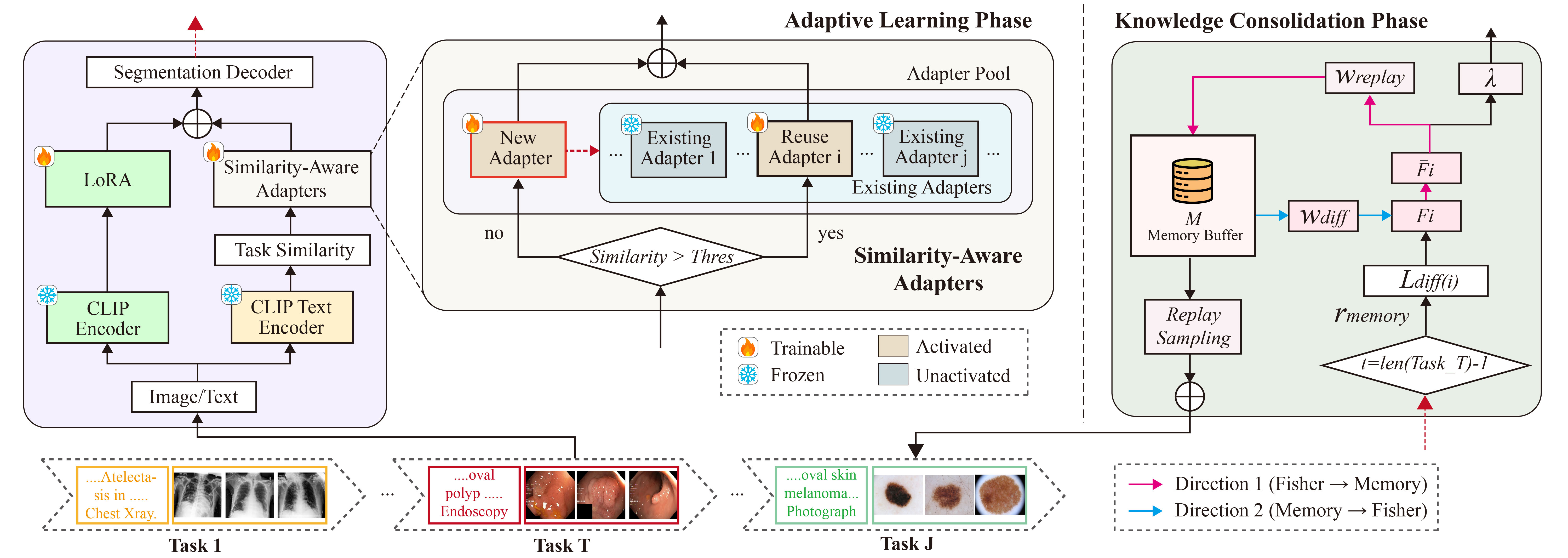} 
\caption{Pipeline of MedPEFT-CL. Our method creates a reinforcement cycle between knowledge consolidation and new task learning. The consolidation phase prevents catastrophic forgetting through bidirectional Fisher-memory feedback, while the adaptive learning phase enables efficient adaptation to new medical segmentation tasks via semantic similarity-based adapter allocation. This cycle ensures that challenging samples enhance importance computation while preserving cross-modal knowledge.}
\label{fig:overall}
\end{figure*}


In medical continual learning, we encounter a sequence of segmentation tasks $\mathcal{T} = \{T^{(1)}, T^{(2)}, \ldots, T^{(N)}\}$ where each task $T^{(t)}$ involves segmenting anatomical structures using vision-language prompts. Each task is defined by dataset $\mathcal{D}^{(t)} = \{(\mathbf{x}_i, \mathbf{y}_i, p^{(t)})\}$, where $\mathbf{x}_i$ are medical images, $\mathbf{y}_i$ are segmentation masks, and $p^{(t)}$ is the medical prompt.
The challenge is learning new tasks while preserving performance on previous tasks, addressing catastrophic forgetting with parameter efficiency for clinical deployment. Our objective is to minimize:
\begin{equation}
\min_\theta \sum_{k=1}^{t} \mathbb{E}_{(\mathbf{x}, \mathbf{y}) \sim \mathcal{D}^{(k)}} [\ell(f_\theta(\mathbf{x}, p^{(k)}), \mathbf{y})]
\end{equation}
where $\theta$ represents model parameters, $f_\theta(\mathbf{x}, p^{(k)})$ is the model prediction for image $\mathbf{x}$ with prompt $p^{(k)}$, $\ell$ is the segmentation loss, and the summation ensures performance across all learned tasks $k=1$ to $t$.

\subsection{Framework Overview}  

In this paper, we present MedPEFT-CL, a parameter-efficient continual learning framework that addresses catastrophic forgetting in medical vision-language segmentation. Our approach employs a dual-phase architecture that combines adaptive learning with bidirectional knowledge consolidation on the CLIPSeg backbone, leveraging pre-trained vision-language representations while introducing specialized medical-domain adaptations, as shown in Figure~\ref{fig:overall}.

The Adaptive Learning Phase enables efficient adaptation to new medical segmentation tasks through semantic similarity-based adapter allocation and parameter-efficient fine-tuning, while preserving cross-modal knowledge. Our approach trains only 0.24-0.39M parameters per task compared to full model fine-tuning, ensuring computational scalability for clinical deployment.
The Knowledge Consolidation Phase preserves previously acquired medical knowledge during continual learning. This phase prevents catastrophic forgetting through bidirectional Fisher-memory feedback that maintains computational efficiency while reinforcing critical knowledge.

These phases create a reinforcement cycle where consolidation guides memory replay prioritization and influences subsequent task training, while new task acquisition identifies challenging samples that enhance importance computation through difficulty-based weighting. This bidirectional coordination ensures critical knowledge receives prioritized rehearsal over time.


\subsection{Adaptive Learning Phase}

\subsubsection{Medical-Informed Prompts}

We employ clinically-informed prompts adapted from ~\cite{poudel2024exploringtransferlearningmedical} that integrate comprehensive medical domain knowledge for optimal vision-language alignment. Our medical prompts incorporate: (1) pathological explanations defining clinical significance, (2) patient demographics (age, gender), (3) imaging parameters (quality, cardiac cycle phase), (4) geometric properties (shape, size), (5) quantitative attributes (object count), (6) spatial positioning within anatomical structures, and (7) multi-condition awareness referencing co-occurring pathologies.

Example prompts include: \textit{``Left ventricular cavity in two-chamber view of the heart at end of the diastole cycle of a 56-year-old female with good image quality''} and \textit{``One small white round polyp which is a projecting growth of tissue located in right of the image''}. This comprehensive medical contextualization enables effective task similarity computation through discriminative prompt embeddings.

\subsubsection{Task Similarity-Based Adapter}

We propose an intelligent adapter selection mechanism that leverages semantic similarity between medical task prompts to enable strategic adapter reuse, resulting in substantial parameter reduction across related medical tasks.

For a new task $T^{(n)}$ with prompt $p_n$, we compute semantic similarity with previously learned tasks using CLIPSeg's text encoder:
\begin{equation}
s_{n,k} = \frac{f_{\text{text}}(p_n) \cdot f_{\text{text}}(p_k)}{||f_{\text{text}}(p_n)|| \cdot ||f_{\text{text}}(p_k)||}
\end{equation}
where $f_{\text{text}}(\cdot)$ represents the CLIPSeg text encoder and $p_k$ is the representative prompt for task $T^{(k)}$.

The adapter reuse decision follows: if $\max_k s_{n,k} > \tau$ (where $\tau = 0.75$), we reuse the adapter from the most similar task $k^* = \arg\max_k s_{n,k}$. Otherwise, we create a new adapter. This mechanism enables parameter sharing across semantically related medical segmentation tasks while maintaining task-specific adaptations for distinct anatomical regions.

\subsubsection{Bi-Modal Parameter-Efficient Fine-Tuning}

We implement cross-modal PEFT adapters that simultaneously fine-tune both visual and textual modalities within CLIPSeg through low-rank adaptation~\cite{hu2021lora}.

Our bi-modal PEFT applies LoRA to CLIPSeg's transformer layers while freezing pre-trained parameters:
\begin{equation}
W' = W + \Delta W = W + \frac{\alpha}{r}BA
\end{equation}
where $B \in \mathbb{R}^{d \times r}$, $A \in \mathbb{R}^{r \times k}$ with rank $r = 8$ and scaling factor $\alpha = 16$.

This approach requires only 0.24-0.39M trainable parameters per task compared to CLIPSeg's 150M parameters. This is achieved using lightweight task adapters that fine-tune both visual and textual representations while preserving the foundational knowledge from pre-training.


\subsection{Knowledge Consolidation Phase}

\subsubsection{Memory Buffer Construction}

We maintain a memory buffer using a fixed ratio $r_{\text{memory}} = 0.15$ to store the most difficult samples from each task. For each training sample, difficulty is measured by cross-entropy loss:
\begin{equation}
\mathcal{L}_{\text{diff}}(i) = \text{CrossEntropy}(f(\mathbf{x}_i), \mathbf{y}_i)
\end{equation}
The top-$\lfloor |\mathcal{D}^{(t)}| \cdot r_{\text{memory}} \rfloor$ samples with highest loss are selected for storage in $\mathcal{D}_{\text{hard}}^{(t)}$, where the floor function ensures integer sample counts.

The memory buffer stores three components per task:
\begin{equation}
\mathcal{M}^{(t)} = \{(\mathcal{D}_{\text{hard}}^{(t)}, \mathbf{L}_{\text{diff}}^{(t)}, \mathbf{S}_{\text{fisher}}^{(t)})\}
\end{equation}
where $\mathbf{L}_{\text{diff}}^{(t)}$ contains difficulty scores and $\mathbf{S}_{\text{fisher}}^{(t)}$ stores Fisher-adjusted weights for replay sampling.

\subsubsection{Bidirectional Feedback Mechanism}

We implement a bidirectional feedback mechanism between Fisher information and memory replay to preserve critical medical knowledge across tasks~\cite{Kirkpatrick2017}.

\paragraph{Direction 1 (Fisher $\rightarrow$ Memory): Importance-Weighted Replay}

Tasks with higher Fisher importance receive boosted replay weights during memory sampling:
\begin{equation}
w_{\text{replay}}^{(t)} = w_{\text{base}}^{(t)} \cdot (1 + \alpha \cdot \bar{F}^{(t)})
\end{equation}
The average Fisher importance for task $t$ is computed as:
\begin{equation}
\bar{F}^{(t)} = \frac{1}{|\Theta|} \sum_{i=1}^{|\Theta|} \mathbb{E}_{\theta_i} [F_i^{(t)}]
\end{equation}
where $|\Theta|$ is the total number of parameters, $F_i^{(t)}$ is the Fisher information for parameter $\theta_i$ in task $t$, and $\mathbb{E}_{\theta_i} [F_i^{(t)}]$ represents the expected Fisher value over parameter $\theta_i$. The base replay weight $w_{\text{base}}^{(t)} = \frac{1}{T_{\text{current}} - t}$ provides temporal decay with $\alpha = 0.5$. These weights determine the sampling probability of each task's stored samples during replay.

\paragraph{Direction 2 (Memory $\rightarrow$ Fisher): Difficulty-Weighted Fisher Information}

Fisher information computation uses stored difficult samples from memory instead of the full dataset, and applies difficulty-based weighting:
\begin{equation}
F_i^{(t)} = \mathbb{E}_{(\mathbf{x},\mathbf{y}) \in \mathcal{D}_{\text{hard}}^{(t)}} \left[ w_{\text{diff}}(\mathbf{x}) \cdot \left( \frac{\partial \mathcal{L}(\mathbf{x}, \mathbf{y})}{\partial \theta_i} \right)^2 \right]
\end{equation}
where $\mathbb{E}_{(\mathbf{x},\mathbf{y}) \in \mathcal{D}_{\text{hard}}^{(t)}}$ denotes expectation over difficult samples stored in memory, $\frac{\partial \mathcal{L}(\mathbf{x}, \mathbf{y})}{\partial \theta_i}$ is the gradient of loss $\mathcal{L}$ with respect to parameter $\theta_i$, and $w_{\text{diff}}(\mathbf{x}) = 1 + \frac{\mathcal{L}_{\text{diff}}(\mathbf{x})}{\max(\mathcal{L}_{\text{diff}})}$ amplifies Fisher information for more difficult samples. This memory-driven approach focuses Fisher computation on challenging cases.

This bidirectional feedback creates a reinforcement loop where Fisher importance and memory replay mutually strengthen each other. Tasks with high Fisher importance get more replay opportunities, which generates more gradient updates and maintains their Fisher values over time. Conversely, the memory buffer stores difficult samples that produce stronger Fisher information, leading to better parameter protection and more informed replay weighting.

\subsubsection{Replay Training}

Building upon the bidirectional feedback mechanism, we implement a structured replay training protocol that actively integrates stored difficult samples into new task learning. During training for task $t$, the system constructs mixed batches controlled by the replay ratio parameter $r_{\text{replay}}$, where $(1-r_{\text{replay}}) \times 100\%$ of each batch comprises current task samples and $r_{\text{replay}} \times 100\%$ consists of strategically selected replay samples from the memory buffer. The replay sampling process implements the Fisher-weighted importance scores established in Equation (6), where tasks with higher Fisher importance receive proportionally more replay opportunities during training.

The training procedure computes gradients jointly across both current and replay samples through a balanced loss formulation that integrates segmentation accuracy, shape preservation, and knowledge retention:
\begin{equation}
\mathcal{L}_{\text{total}} = \mathcal{L}_{\text{seg}} + \mathcal{L}_{\text{dice}} + \lambda_{\text{ewc}} \cdot \mathcal{L}_{\text{EWC}}
\end{equation}
The EWC loss constrains parameter drift from previously learned tasks:
\begin{equation}
\mathcal{L}_{\text{EWC}} = \sum_{t'<t} \sum_i F_i^{(t')} (\theta_i - \theta_i^{(t')})^2
\end{equation}
where $t'$ represents previous tasks ($t' < t$), $i$ indexes model parameters, $F_i^{(t')}$ is the Fisher information for parameter $\theta_i$ from task $t'$, $\theta_i$ is the current parameter value, $\theta_i^{(t')}$ is the parameter value after completing task $t'$, and $\lambda_{\text{ewc}} = 500$ provides EWC regularization strength.
 This replay training ensures that challenging medical cases from previous anatomical domains remain actively integrated into the learning process.


\section{Experiments}
\label{sec:Experiments}

\subsection{Datasets and Evaluation Metric}

\noindent\textbf{Datasets.}
We evaluate our framework on multiple medical segmentation datasets across diverse modalities (Table~\ref{tab:medical_datasets}): endoscopic polyp segmentation~\cite{jha2020kvasir,bernal2015wm,silva2014toward,tajbakhsh2016automated,vazquez2017benchmark}, dermatological imaging~\cite{gutman2016skin,cassidy2022dfuc2022}, ultrasound~\cite{leclerc2019deep,al2020dataset}, and chest X-ray~\cite{irvin2019chexpert}. These datasets span different anatomical regions (colon, skin, heart, breast, chest) and imaging characteristics (color endoscopy, photography, grayscale ultrasound, X-ray), providing comprehensive evaluation scenarios for continual learning performance. This multi-modal collection enables robust assessment of our approach across varying visual characteristics and clinical contexts.

\begin{table}[H]
\centering
\caption{Overview of medical image segmentation datasets organized by imaging modality and anatomical region.}
\fontsize{9pt}{12pt}\selectfont
\label{tab:medical_datasets}
\begin{tabular*}{\columnwidth}{@{\extracolsep{\fill}}llll}
\hline
Modality & Organ & Name & \# train/val/test \\
\hline
\multirow{6}{*}{Endoscopy} & \multirow{6}{*}{Colon} & Kvasir-SEG~\cite{jha2020kvasir} & 800/100/100 \\
 & & ClinicDB~\cite{bernal2015wm} & 490/61/61 \\
 & & ETIS~\cite{silva2014toward} & 157/20/19 \\
 & & ColonDB~\cite{tajbakhsh2016automated} & 304/38/38 \\
 & & CVC300~\cite{vazquez2017benchmark} & 48/6/6 \\
\hline
\multirow{2}{*}{Photography} & Skin & ISIC 2016~\cite{gutman2016skin} & 810/90/379 \\
 & Foot & DFU 2022~\cite{cassidy2022dfuc2022} & 1600/200/200 \\
\hline
\multirow{2}{*}{Ultrasound} & Heart & CAMUS~\cite{leclerc2019deep} & 4800/600/600 \\
 & Breast & BUSI~\cite{al2020dataset} & 624/78/78 \\
\hline
X-Ray & Chest & CheXlocalize~\cite{irvin2019chexpert} & 1279/446/452 \\
\hline
\end{tabular*}
\end{table}


\begin{table*}[t]
\centering
\caption{\textit{\textbf{Mixed Task Sequence: Kvasir$\rightarrow$CheX$\rightarrow$Colon$\rightarrow$BUSI$\rightarrow$Clinic$\rightarrow$ISIC$\rightarrow$ETIS$\rightarrow$Camus$\rightarrow$CVC300$\rightarrow$DFU}}.Performance comparison for mixed task sequences. "CLIPSeg Individual" represents upper bound performance when each task is trained independently; "CLIPSeg Sequential" represents naive sequential training without continual learning techniques. Avg Dice shows segmentation performance; Avg FR (\%) shows forgetting rate (lower is better). All results are averaged over 3 independent runs. \textbf{Bold} indicates best performance among continual learning methods.}
\label{tab:mixed_task_sequence}
\fontsize{9pt}{12pt}\selectfont
\resizebox{\textwidth}{!}{%
\begin{tabular}{p{0.8cm}|l|cccccccccc|c}
\hline
Metric & Method & Kvasir & CheX & Colon & BUSI & Clinic & ISIC & ETIS & Camus & CVC300 & DFU & Avg \\
\hline
\multirow{8}{*}{\begin{tabular}{c}Avg \\ Dice\end{tabular}} & CLIPSeg Individual& 91.39 & 62.57 & 78.07 & 69.31 & 85.69 & 91.94 & 75.93 & 86.09 & 92.90 & 76.31 & 81.02 \\
\cline{2-13}
 & CLIPSeg Sequential & 67.44 & 42.47 & 59.12 & 55.64 & 68.60 & 90.54 & 62.39 & 74.66 & 90.92 & 72.28 & 68.41 \\
 & EWC (2017)\cite{Kirkpatrick2017} & 76.50 & 48.47 & 71.01 & 56.64 & 77.36 & 85.01 & 68.07 & 74.51 & \textbf{93.50} & 72.88 & 72.40 \\

 & DER (2021)\cite{Buzzega2021} & 78.95 & 51.23 & 66.44 & 61.87 & 72.15 & 89.92 & \textbf{69.81} & 78.94 & 92.18 & 71.92 & 73.34 \\
 & VR-LwF (2022)\cite{VR-LwF2022} & 75.67 & 48.91 & 69.55 & 60.18 & 76.83 & 89.15 & 67.22 & 77.63 & 92.01 & 74.18 & 73.13 \\
 & Low-Rank MoE (2024)\cite{Che_LowRank_MICCAI2024}  & 73.82 & 46.15 & 67.89 & 58.92 & 74.28 & 88.76 & 65.43 & 76.88 & 91.67 & \textbf{73.55} & 71.74 \\
 & MoE-Adapter (2024)\cite{MoE-Adapter2024} & 79.28 & 53.17 & 68.29 & 65.48 & 72.96 & 90.82 & 68.89 & 78.11 & 91.30 & 73.17 & 74.15 \\
 & \cellcolor{pink!30}\textbf{Ours} & \cellcolor{pink!30}\textbf{85.85} & \cellcolor{pink!30}\textbf{68.72} & \cellcolor{pink!30}\textbf{75.27} & \cellcolor{pink!30}\textbf{70.67} & \cellcolor{pink!30}\textbf{81.46} & \cellcolor{pink!30}\textbf{91.77} & \cellcolor{pink!30}64.04 & \cellcolor{pink!30}\textbf{80.18} & \cellcolor{pink!30}91.61 & \cellcolor{pink!30}72.28 & \cellcolor{pink!30}\textbf{78.19} \\
\hline
\multirow{7}{*}{\begin{tabular}{c}Avg \\ FR \\ (\%)\end{tabular}} & CLIPSeg Sequential & 17.43 & 24.87 & 14.55 & 23.84 & 13.00 & 1.81 & 6.08 & 3.54 & 2.00 & - & 11.90 \\
 & EWC (2017)\cite{Kirkpatrick2017} & 10.14 & 16.77 & 6.14 & 18.92 & 9.74 & 8.32 & 3.81 & 9.55 & 1.05 & - & 9.38 \\
 & DER (2021)\cite{Buzzega2021} & 8.92 & 13.45 & 8.28 & 14.18 & 12.15 & 3.54 & \textbf{2.95} & 6.41 & 1.88 & - & 7.97 \\
 & VR-LwF (2022)\cite{VR-LwF2022} & 11.35 & 17.81 & 7.63 & 15.46 & 10.28 & 3.89 & 4.67 & 7.12 & 1.24 & - & 8.83 \\
  & Low-Rank MoE (2024)\cite{Che_LowRank_MICCAI2024} & 12.68 & 19.23 & 8.87 & 16.74 & 11.92 & 4.26 & 5.32 & 7.83 & 1.89 & - & 9.86 \\
 & MoE-Adapter (2024)\cite{MoE-Adapter2024} & 6.90 & 8.79 & 9.03 & 8.62 & 11.41 & 2.31 & 3.76 & 2.76 & 3.00 & - & 6.21 \\
 & \cellcolor{pink!30}\textbf{Ours} & \cellcolor{pink!30}\textbf{0.77} & \cellcolor{pink!30}\textbf{0.90} & \cellcolor{pink!30}\textbf{0.26} & \cellcolor{pink!30}\textbf{4.53} & \cellcolor{pink!30}\textbf{1.34} & \cellcolor{pink!30}\textbf{1.16} & \cellcolor{pink!30}7.29 & \cellcolor{pink!30}\textbf{1.42} & \cellcolor{pink!30}\textbf{0.90} & \cellcolor{pink!30}- & \cellcolor{pink!30}\textbf{2.23} \\
\hline
\end{tabular}%
}
\end{table*}

\begin{figure*}[t]
\centering
\includegraphics[width=\textwidth]{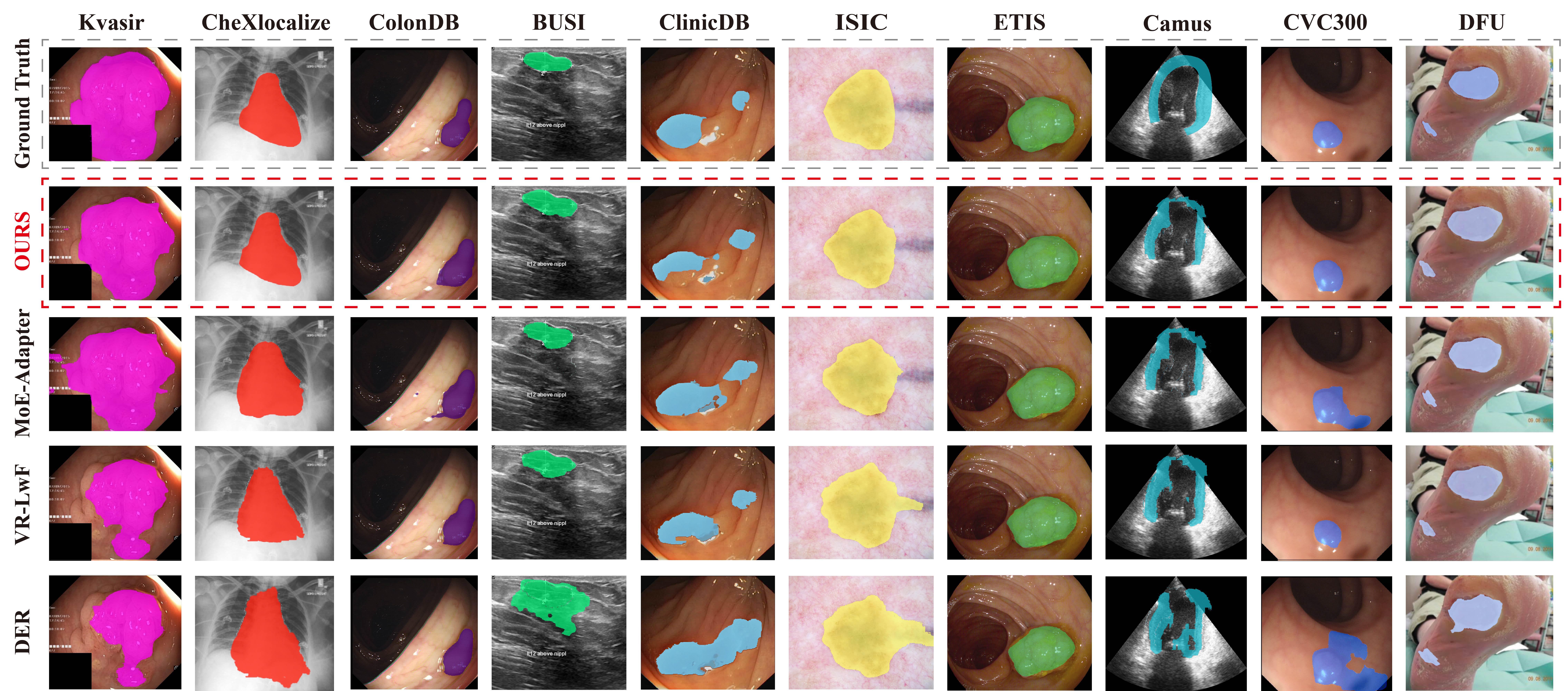} 
\caption{\textit{\textbf{Mixed Task Sequence: Kvasir$\rightarrow$CheX$\rightarrow$Colon$\rightarrow$BUSI$\rightarrow$Clinic$\rightarrow$ISIC$\rightarrow$ETIS$\rightarrow$Camus$\rightarrow$CVC300$\rightarrow$DFU}}.Visualization for mixed task sequences. Segmentation results from top row to bottom: Ground-truth, Ours, Moe-adapter, VR-LwF, DER.}
\label{fig:Mixed Task Sequence}
\end{figure*}

\vspace{0.3cm}
\noindent\textbf{Metrics.}
We assess continual learning performance using two primary metrics:

(1) \textbf{Average Dice Coefficient} measuring segmentation accuracy across all tasks, computed as
\begin{equation}
\text{Avg Dice} = \frac{1}{T}\sum_{i=1}^{T} \frac{2|P_i \cap G_i|}{|P_i| + |G_i|}
\end{equation}
where $P_i$ and $G_i$ are the predicted and ground truth segmentation masks for task $i$, and $T$ is the total number of tasks;

(2) \textbf{Average Forgetting Rate} quantifying knowledge retention, calculated as
\begin{equation}
\text{FR} = \frac{1}{T-1}\sum_{i=1}^{T-1}\frac{\text{Dice}_i^{peak} - \text{Dice}_i^{final}}{\text{Dice}_i^{peak}} \times 100\%
\end{equation}
where $\text{Dice}_i^{peak}$ is the peak performance achieved on task $i$ during training and $\text{Dice}_i^{final}$ is the final performance after learning all subsequent tasks. 
We introduce this peak-based formulation, which is more rigorous than Lopez-Paz's initial performance baseline approach~\cite{lopez2017gradient} for medical scenarios. Our peak-based formulation ensures clinical reliability by measuring degradation from validated maximum diagnostic capability rather than suboptimal initial training performance. This approach is essential for clinical deployment where accurate assessment of diagnostic capability loss directly impacts patient safety and decision-making.

\vspace{0.5cm}
\noindent\textbf{Implementation Details.} 
Our framework uses CLIPSeg-RD64 backbone with training conducted on 2$\times$A100 GPUs. Images are resized to 352$\times$352 with AdamW optimizer (lr=$8 \times 10^{-4}$, weight decay=$8 \times 10^{-5}$), batch size 16, and maximum 100 epochs per task with early stopping. Task similarity threshold is set to 0.75. All hyperparameters including LoRA adaptation ($r=8$, $\alpha=16$), memory ratio ($r_{\text{memory}}$ = 0.15), replay ratio ($r_{\text{replay}}$ = 0.4) and EWC regularization ($\lambda=500$) are validated as the optimal configuration through comprehensive analysis in Table~\ref{tab:Parameter Efficiency Analysis}. All results are averaged over three runs with random seed 43.


\subsection{Comparison with State-of-the-Art Methods}

We evaluate our framework against state-of-the-art continual learning methods to demonstrate its effectiveness in mitigating catastrophic forgetting for medical image segmentation. To comprehensively assess robustness across different clinical deployment scenarios, we design three distinct evaluation protocols: mixed sequences combining diverse medical imaging modalities and anatomical regions to simulate real-world clinical workflows, heterogeneous sequences testing cross-domain knowledge transfer between fundamentally different imaging types, and homogeneous sequences focusing on related anatomical structures to evaluate fine-grained knowledge retention. Each sequence tests different aspects of continual learning capability essential for practical medical AI deployment, where models must adapt to new imaging protocols while maintaining diagnostic accuracy on previously learned tasks. Our experiments reveal that catastrophic forgetting is a critical issue in medical vision-language models in all sequences, as evidenced by CLIPSeg Sequential's dramatic performance degradation compared to CLIPSeg Individual (where each task is trained independently as an upper bound). We compare against both traditional continual learning baselines such as EWC~\cite{Kirkpatrick2017}, as well as emerging methods including DER~\cite{Buzzega2021}, VR-LwF~\cite{VR-LwF2022}, low-rank MoE~\cite{Che_LowRank_MICCAI2024} and MoE-Adapter~\cite{MoE-Adapter2024}.

\paragraph{Comparison on Mixed Task Sequences}

Table~\ref{tab:mixed_task_sequence} and Figure~\ref{fig:Mixed Task Sequence} evaluates our method on mixed task sequences combining diverse medical imaging modalities to simulate realistic clinical scenarios. The catastrophic forgetting problem is clearly evident: CLIPSeg Sequential suffers severe performance degradation with 11.90\% forgetting rate compared to the 81.02\% upper bound from independent training. Our MedPEFT-CL achieves superior performance with 78.19\% average Dice score, outperforming the best baseline MoE-Adapter (74.15\%) by 4.04 percentage points. Most critically, our method demonstrates forgetting mitigation with only 2.23\% average forgetting rate, representing a 64\% reduction compared to MoE-Adapter's 6.21\%.

\paragraph{Comparison on Heterogeneous Task Sequences}

Table~\ref{tab:heterogeneous_task_sequence} examines cross-domain knowledge transfer across fundamentally different imaging modalities. This challenging scenario tests the model's ability to retain knowledge when transitioning between diverse medical domains such as endoscopy, dermatology, radiology, and ultrasound. CLIPSeg Sequential exhibits even more severe catastrophic forgetting with substantial performance drops, particularly on CheX (62.57\% to 25.25\%) and Kvasir (91.39\% to 58.34\%). Our MedPEFT-CL demonstrates robust cross-domain adaptation, achieving superior performance across most tasks while maintaining exceptionally low forgetting rates. Notably, our method achieves only 0.66\% forgetting on CheX and 0.57\% on ISIC, compared to MoE-Adapter's 11.49\% and 7.01\% respectively. 

\begin{table}[H]
\centering
\caption{\textit{\textbf{Heterogeneous Task Sequence: Kvasir$\rightarrow$ISIC$\rightarrow$CheX \\
$\rightarrow$BUSI$\rightarrow$Camus$\rightarrow$DFU}}.Performance comparison for heterogeneous task sequences. \textbf{Bold} indicates best performance.}
\label{tab:heterogeneous_task_sequence}
\fontsize{9pt}{13pt}\selectfont
\resizebox{\columnwidth}{!}{%
\begin{tabular}{p{0.7cm}|p{2.4cm}|>{\centering\arraybackslash}p{0.7cm}>{\centering\arraybackslash}p{0.7cm}>{\centering\arraybackslash}p{0.7cm}>{\centering\arraybackslash}p{0.7cm}>{\centering\arraybackslash}p{0.7cm}>{\centering\arraybackslash}p{0.7cm}}
\hline
Met. & Method & Kvasir & ISIC & CheX & BUSI & Camus & DFU \\
\hline
\multirow{8}{*}{\begin{tabular}{c}Avg \\ Dice\end{tabular}} & CLIPSeg Ind. & 91.39 & 91.94 & 62.57 & 69.31 & 91.09 & 76.13 \\
\cline{2-8}
 & CLIPSeg Seq. & 58.34 & 84.26 & 25.25 & 62.37 & 79.80 & 72.12 \\
 & EWC & 63.02 & 86.39 & 38.36 & 63.07 & 84.12 & 73.04 \\
 & DER & 73.22 & 88.45 & 53.91 & 66.18 & \textbf{85.78} & \textbf{74.12} \\
 & VR-LwF & 67.89 & 86.78 & 45.33 & 63.95 & 83.44 & 73.45 \\
 & Low-Rank MoE & 65.45 & 87.12 & 42.18 & 64.23 & 82.95 & 73.67 \\
 & MoE-Adapter & 71.38 & 84.87 & 50.05 & 69.16 & 79.39 & 73.02 \\
 & \cellcolor{pink!30}\textbf{Ours} & \cellcolor{pink!30}\textbf{81.78} & \cellcolor{pink!30}\textbf{92.15} & \cellcolor{pink!30}\textbf{69.63} & \cellcolor{pink!30}\textbf{70.69} & \cellcolor{pink!30}81.66 & \cellcolor{pink!30}72.28 \\
\hline
\multirow{7}{*}{\begin{tabular}{c}Avg \\ FR \\ (\%)\end{tabular}} & CLIPSeg Seq. & 27.22 & 8.30 & 41.69 & 15.68 & 4.51 & - \\
 & EWC & 24.12 & 7.07 & 29.56 & 15.42 & 1.68 & - \\
 & DER & 13.78 & 4.23 & 15.34 & 12.45 & 1.95 & - \\
 & VR-LwF & 18.34 & 5.67 & 19.88 & 13.67 & 2.44 & - \\
 & Low-Rank MoE & 21.67 & 6.45 & 25.12 & 14.78 & 2.89 & - \\
 & MoE-Adapter & 15.11 & 7.01 & 11.49 & 8.93 & 3.12 & - \\
 & \cellcolor{pink!30}\textbf{Ours} & \cellcolor{pink!30}\textbf{3.78} & \cellcolor{pink!30}\textbf{0.57} & \cellcolor{pink!30}\textbf{0.66} & \cellcolor{pink!30}\textbf{7.82} & \cellcolor{pink!30}\textbf{0.21} & \cellcolor{pink!30}- \\
\hline
\end{tabular}%
}
\end{table}

\paragraph{Comparison on Homogeneous Task Sequences}

Table~\ref{tab:homogeneous_task_sequence_homo1} evaluates performance on homogeneous task sequences focusing on related anatomical structures within the gastrointestinal domain. This scenario tests fine-grained knowledge retention when learning semantically similar medical tasks that share visual and diagnostic characteristics. Even within this related domain, CLIPSeg Sequential shows noticeable performance degradation across all tasks. Our MedPEFT-CL achieves competitive performance while demonstrating superior forgetting mitigation, with particularly strong results on CVC (90.90\%) and Colon (77.67\%) tasks. Most importantly, our method maintains exceptionally low forgetting rates across most tasks, achieving only 1.19\% on Colon and 1.96\% on CVC, substantially outperforming baselines. 

\begin{table}[H]
\centering
\caption{\textit{\textbf{Homogeneous Task Sequence: Clinic$\rightarrow$ETIS$\rightarrow$CVC\\
$\rightarrow$Colon$\rightarrow$Kvasir}}. Performance for homogeneous task sequences.\textbf{Bold} indicates best performance.}
\label{tab:homogeneous_task_sequence_homo1}
\fontsize{10pt}{13pt}\selectfont
\resizebox{\columnwidth}{!}{%
\begin{tabular}{p{0.8cm}|p{2.3cm}|>{\centering\arraybackslash}p{0.9cm}>{\centering\arraybackslash}p{0.9cm}>{\centering\arraybackslash}p{0.9cm}>{\centering\arraybackslash}p{0.9cm}>{\centering\arraybackslash}p{0.9cm}}
\hline
Metric & Method & Clinic & ETIS & CVC & Colon & Kvasir \\
\hline
\multirow{8}{*}{\begin{tabular}{c}Avg \\ Dice\end{tabular}} & CLIPSeg Ind. & 91.69 & 75.93 & 88.90 & 78.07 & 91.39 \\
\cline{2-7}
 & CLIPSeg Seq. & 80.80 & 61.38 & 83.24 & 71.57 & 85.98 \\
 & EWC & 82.05 & 74.13 & 85.44 & 69.86 & 88.40 \\
 & DER & \textbf{82.35} & \textbf{74.85} & 85.80 & 70.45 & \textbf{88.75} \\
 & VR-LwF & 81.90 & 73.65 & 85.20 & 70.80 & 88.25 \\
 & Low-Rank MoE & 81.45 & 72.80 & 84.90 & 70.25 & 87.85 \\
 & MoE-Adapter & 79.71 & 66.91 & 88.38 & 71.68 & 87.14 \\
 & \cellcolor{pink!30}\textbf{Ours} & \cellcolor{pink!30}82.21 & \cellcolor{pink!30}69.99 & \cellcolor{pink!30}\textbf{90.90} & \cellcolor{pink!30}\textbf{77.67} & \cellcolor{pink!30}87.08 \\
\hline
\multirow{7}{*}{\begin{tabular}{c}Avg \\ FR \\ (\%)\end{tabular}} & CLIPSeg Seq. & 5.00 & 11.58 & 8.60 & 8.61 & - \\
 & EWC & 4.06 & \textbf{0.94} & 9.08 & 10.04 & - \\
 & DER & 4.05 & 1.85 & 5.45 & 8.85 & - \\
 & VR-LwF & 4.35 & 2.95 & 5.45 & 8.40 & - \\
 & Low-Rank MoE & 4.58 & 3.42 & 6.20 & 8.95 & - \\
 & MoE-Adapter & 4.14 & 5.29 & 8.25 & 6.69 & - \\
 & \cellcolor{pink!30}\textbf{Ours} & \cellcolor{pink!30}\textbf{1.73} & \cellcolor{pink!30}4.34 & \cellcolor{pink!30}\textbf{1.96} & \cellcolor{pink!30}\textbf{1.19} & \cellcolor{pink!30}- \\
\hline
\end{tabular}%
}
\end{table}

\paragraph{Computational Cost}
Table \ref{tab:method_comparison} compares computational efficiency and performance across continual learning methods. We compare our approach against several baseline methods including EWC~\cite{Kirkpatrick2017}, Learning without Forgetting (LwF)\cite{Li2017}, PackNet\cite{mallya2018packnet}, Progressive Neural Networks~\cite{rusu2016progressive}, Dark Experience Replay (DER)\cite{Buzzega2021}, VR-LwF\cite{VR-LwF2022}, Low-Rank MoE~\cite{Che_LowRank_MICCAI2024}, and MoE-Adapters~\cite{MoE-Adapter2024}. 

\begin{table}[H]
\centering
\caption{Comparison of computational cost during training. Avg DICE and Avg FR are computed across all evaluated task sequences (mixed, heterogeneous, and homogeneous).}
\label{tab:method_comparison}
\resizebox{\columnwidth}{!}{%
\fontsize{10pt}{12.5pt}\selectfont
\begin{tabular}{l|cc|ccc}
\hline
\multirow{3}{*}{Method} & \multicolumn{2}{c|}{Performance} & \multicolumn{3}{c}{Efficiency} \\
\cline{2-6}
 & Avg & Avg FR & Trainable & GPU & Time \\
 & DICE & (\%) & Params (M) & (MiB) & (s/it) \\
\hline
CLIPSeg Ind. & 81.02 & - & 150 & - & - \\
CLIPSeg Seq. & 67.71 & 15.36 & 150 & 32172 & 1.50 \\
EWC & 71.37 & 10.73 & 150 & 32180 & 1.52 \\
LwF & 70.86 & 11.48 & 149.6 & 32172 & 1.54 \\
PackNet & 74.21 & 8.32 & 75 & 16086 & 0.95 \\
Progressive NN & 73.12 & 7.89 & 375 & 80430 & 3.75 \\
DER & 74.59 & 6.58 & 155 & 33276 & 1.68 \\
VR-LwF & 73.69 & 8.18 & 149.6 & 32236 & 1.51 \\
Low-Rank MoE & 72.86 & 10.48 & 2.3 & 15,247 & 1.07 \\
MoE-Adapters & 75.81 & 5.13 & 59.8 & 19898 & 1.37 \\
\cellcolor{pink!30}Ours & \cellcolor{pink!30}78.84 & \cellcolor{pink!30}1.91& \cellcolor{pink!30}0.24-0.39 & \cellcolor{pink!30}22358 & \cellcolor{pink!30}1.58 \\
\hline
\end{tabular}%
}
\end{table}

Our approach achieves the optimal efficiency-performance trade-off, using only 0.24--0.39M trainable parameters (up to 1000$\times$ fewer than Progressive NN) while maintaining a 78.84\% average DICE score and the lowest forgetting rate of 1.91\%. Notably, our method outperforms the most efficient baseline MoE-Adapter (75.81\% DICE, 5.13\% forgetting) while using significantly fewer parameters. With moderate GPU memory requirements (22,358 MiB) and training time (1.58 s/iteration), our method demonstrates that exceptional knowledge retention and segmentation quality can be achieved with minimal computational overhead.


\subsection{Ablation Study}

\paragraph{Semantic Adapter Allocation Analysis}
The semantic adapter allocation analysis shows adapter allocation across similarity thresholds for medical vision-language tasks (Table \ref{tab:semantic_adapter_allocation}). The 0.75 threshold achieves optimal performance with 60\% new adapters and 40\% reused adapters, delivering the highest Dice score (78.84\%) and lowest forgetting rate (1.91\%). This threshold is most aligned with the dataset's natural characteristics, where anatomically related tasks (polyp segmentation) effectively share adapters while cross-modal tasks (endoscopy vs. ultrasound vs. X-ray) require specialized adaptations. Lower thresholds lead to excessive adapter reuse and performance degradation, while higher thresholds prevent effective knowledge transfer by creating too many unnecessary adapters.

\begin{table}[H]
\centering
\caption{Semantic adapter allocation analysis across different similarity thresholds showing adapter reuse decisions, performance outcomes, and forgetting rates. \textbf{Bold} indicates best performance.}
\label{tab:semantic_adapter_allocation}
\resizebox{\columnwidth}{!}{%
\fontsize{10pt}{12.5pt}\selectfont
\begin{tabular}{l|cc|cc}
\hline
\multirow{2}{*}{Threshold} & \multicolumn{2}{c|}{Adapter Allocation} & \multicolumn{2}{c}{Performance Metrics} \\
\cline{2-5}
 & \begin{tabular}{@{}c@{}}New\\Adapters\end{tabular} & \begin{tabular}{@{}c@{}}Reused\\Adapters\end{tabular} & Avg Dice (\%) & Avg FR (\%) \\
\hline
0.3 & 30\% & 70\% & 73.00 & 5.72 \\
0.5 & 45\% & 55\% & 76.52 & 3.28 \\
\textbf{0.75} & \textbf{60\%} & \textbf{40\%} & \textbf{78.84} & \textbf{1.91} \\
0.9 & 65\% & 35\% & 74.52 & 4.22 \\
\hline
\end{tabular}%
}
\end{table}

\paragraph{Parameter Efficiency Analysis}
\label{Parameter Efficiency Analysis}

Our parameter efficiency analysis across nine hyperparameter configurations (Table~\ref{tab:Parameter Efficiency Analysis}) demonstrates optimal trade-offs between performance and efficiency. Config 4 achieves peak performance with 80.11\% average Dice and 1.38\% forgetting rate, while Config 6 offers superior balance (78.84\% Dice, 1.91\% forgetting) with faster training (1.6 vs 2.5 s/it). All configurations maintain exceptional parameter efficiency, using only 0.16--0.26\% (235K--385K) of the $\sim$151M total parameters. While minimal memory configurations like Config 3 improve speed (0.9 s/it), they incur significant performance penalties (69.37\% Dice, 11.28\% forgetting). The analysis reveals that moderate LoRA ranks (r=8-16) with appropriate memory ratio ($r_{\text{memory}}$=0.15-0.20), replay ratio ($r_{\text{replay}}$=0.35-0.50) and regularization strength ($\lambda$=500-2000) provide the best performance-efficiency trade-off for continual learning scenarios.

\begin{table}[H]
\centering
\caption{Hyperparameter analysis results showing the trade-off between memory efficiency, performance, and training time. Avg DICE and Avg FR are computed across all evaluated task sequences (mixed, heterogeneous, and homogeneous).}
\fontsize{8pt}{10pt}\selectfont
\label{tab:Parameter Efficiency Analysis}
\begin{tabular}{>{\centering\arraybackslash}m{0.05cm}|>{\centering\arraybackslash}m{0.25cm}>{\centering\arraybackslash}m{0.25cm}|>{\centering\arraybackslash}m{0.6cm}>{\centering\arraybackslash}m{0.6cm}|>{\centering\arraybackslash}m{0.5cm}|>{\centering\arraybackslash}m{0.7cm}>{\centering\arraybackslash}m{0.45cm}>{\centering\arraybackslash}m{0.45cm}>{\centering\arraybackslash}m{0.45cm}}
\hline
\multirow{2}{*}{\#} & \multicolumn{2}{c|}{LoRA} & \multicolumn{2}{c|}{Memory} & \multirow{2}{*}{$\lambda$} & \multicolumn{4}{c}{Performance \& Time} \\
\cline{2-3} \cline{4-5} \cline{7-10} 
& $r$ & $\alpha$ & \begin{tabular}{@{}c@{}} $r_{\text{memory}}$\end{tabular} & \begin{tabular}{@{}c@{}} $r_{\text{replay}}$\end{tabular} & & \begin{tabular}{@{}c@{}}Train\\Param\end{tabular} & \begin{tabular}{@{}c@{}}Avg\\Dice\end{tabular} & \begin{tabular}{@{}c@{}}Avg\\FR(\%)\end{tabular} & \begin{tabular}{@{}c@{}}Time\\(s/it)\end{tabular} \\
\hline
1 & 8 & 16 & 0.10 & 0.35 & 150 & 245,636 & 75.34 & 5.24 & 1.2 \\
2 & 8 & 16 & 0.15 & 0.35 & 1000 & 245,636 & 76.50 & 4.01 & 1.8 \\
3 & 8 & 16 & 0.05 & 0.2 & 100 & 245,636 & 69.37 & 11.28 & 0.9 \\
4 & 16 & 32 & 0.20 & 0.5 & 2000 & 265,604 & 80.11 & 1.38 & 2.5 \\
5 & 64 & 128 & 0.25 & 0.25 & 100 & 385,412 & 75.38 & 7.06 & 2.3 \\
6 & 8 & 16 & 0.15 & 0.4 & 500 & 245,636 & 78.84 & 1.91 & 1.6 \\
7 & 12 & 24 & 0.25 & 0.25 & 500 & 255,620 & 75.71 & 3.64 & 1.7 \\
8 & 8 & 16 & 0.05 & 0.25 & 1000 & 245,636 & 71.73 & 7.16 & 1.3 \\
9 & 4 & 8 & 0.10 & 0.5 & 150 & 235,652 & 74.74 & 3.95 & 1.5 \\
\hline
\end{tabular}
\end{table}

\paragraph{Module Ablation Analysis}
The module ablation study (Table~\ref{tab:module_ablation}) demonstrates the distinct benefits of each proposed module. From a PEFT baseline of 61.31\% Dice and 20.32\% forgetting, EWC Knowledge Consolidation adds a 5.91\% Dice improvement. Introducing Replay Memory delivers substantial gains (+11.96\% Dice, -13.76\% forgetting), which are further enhanced by Fisher Guidance. The complete framework, incorporating Bidirectional Feedback, achieves a cumulative improvement of +17.53\% Dice and -18.41\% forgetting reduction. This demonstrates the synergistic effect of our integrated components, where each module contributes progressively to improved segmentation accuracy and knowledge retention.

\begin{table}[H]
\centering
\caption{Ablation study of different components. \textbf{Bold} indicates best performance.}
\label{tab:module_ablation}
\resizebox{\columnwidth}{!}{%
\fontsize{10pt}{12.5pt}\selectfont
\begin{tabular}{l|cc|cc}
\hline
\multirow{2}{*}{Components} & \multicolumn{2}{c|}{Avg Dice} & \multicolumn{2}{c}{Avg FR(\%)} \\
\cline{2-5}
 & Score & $\Delta$ & Rate & $\Delta$ \\
\hline
PEFT Baseline & 61.31 & 0 & 20.32 & 0 \\
\ \ +EWC Knowledge Consolidation & 67.22 & +5.91 & 14.01 & -6.31 \\
\ \ +Replay Memory & 73.27 & +11.96 & 6.56 & -13.76 \\
\ \ +Fisher Guidance & 76.24 & +14.93 & 3.38 & -16.94 \\
\cellcolor{pink!30}\ \ \textbf{+Bidirectional Feedback} & \cellcolor{pink!30}\textbf{78.84} & \cellcolor{pink!30}\textbf{+17.53} & \cellcolor{pink!30}\textbf{1.91} & \cellcolor{pink!30}\textbf{-18.41} \\
\hline
\end{tabular}%
}
\end{table}

\section{Conclusion}

We propose MedPEFT-CL, a parameter-efficient continual learning framework that addresses catastrophic forgetting in medical vision-language segmentation through bidirectional feedback mechanisms and semantic adapter allocation. Our computationally lightweight approach consistently outperforms state-of-the-art methods across extensive experiments. The framework provides a robust foundation for evolving medical AI systems that can continuously adapt to new clinical requirements while preserving established diagnostic capabilities, ultimately advancing the development of sustainable healthcare AI solutions.

\clearpage
{
    \small
    \bibliographystyle{ieeenat_fullname}
    \bibliography{main}
}

\end{document}

%% file: preamble.tex
\usepackage{amsthm}
\usepackage{enumitem}
\usepackage{amsmath}
\usepackage{amssymb}
\usepackage{amsfonts}
\usepackage{mathtools}
\usepackage{float}
\usepackage{textcomp}
\usepackage{booktabs}
\usepackage{multirow}
\usepackage{array}
\usepackage{graphicx}
\usepackage{adjustbox}
\usepackage{dsfont}
\usepackage{algorithm}
\usepackage{algorithmic}
\usepackage{tcolorbox}
\usepackage{array}
\usepackage{xcolor}
\usepackage{colortbl}
\usepackage{booktabs}
\usepackage{makecell}
\usepackage{caption}
